\documentclass[10pt,twocolumn,letterpaper]{article}

\usepackage{iccv}
\usepackage{times}
\usepackage{epsfig}
\usepackage{graphicx}
\usepackage{tabulary}
\usepackage{colortbl}
\usepackage{amsmath}
\usepackage{amsthm}
\usepackage{amssymb}
\usepackage{bm}
\usepackage{bbm}
\usepackage{subfig}
\usepackage{multirow}
\usepackage{color}
\usepackage{tabu}
\usepackage{array, eucal}
\usepackage[ruled,vlined]{algorithm2e}
\usepackage[american]{babel}

\usepackage{caption}
\usepackage[pagebackref=true,breaklinks=true,colorlinks,bookmarks=false]{hyperref}

\definecolor{mygray}{gray}{.9}
\newcolumntype{L}{>{\raggedright\arraybackslash}m{3.75cm}}

\newcommand{\mat}[1]{\mbox{\boldsymbol{$#1$}}}

\newcommand{\squishlist}{
	\begin{list}{$\bullet$}
		{ \setlength{\itemsep}{0pt}
			\setlength{\parsep}{2pt}
			\setlength{\topsep}{2pt}
			\setlength{\partopsep}{0pt}
			\setlength{\leftmargin}{1em}
			\setlength{\labelwidth}{1em}
			\setlength{\labelsep}{0.5em} } }
	\newcommand{\squishend}{
\end{list}  }

\graphicspath{{./Figures/}}

\iccvfinalcopy %

\def\iccvPaperID{1411} %

\ificcvfinal\fi
\begin{document}

\title{
Indices Matter: Learning to Index for Deep Image Matting
}

\author{Hao~Lu$^\dagger$ ~ ~ ~ ~ Yutong~Dai$^\dagger$
~ ~ ~ ~
Chunhua~Shen$^\dagger$\thanks{Corresponding author.} ~ ~ ~ ~ Songcen~Xu$^\ddagger$\\
$^\dagger$The University of Adelaide, Australia ~ ~ ~ ~
$^\ddagger$Noah's Ark Lab, Huawei Technologies\\
e-mail: {\tt\small \{$\sf hao.lu,yutong.dai, chunhua.shen\}@adelaide.edu.au$}
}

\maketitle
\thispagestyle{empty}

\begin{abstract}

	We show that existing upsampling operators can be unified with the notion of the index function. This notion is inspired by an observation in the decoding process of deep image matting where indices-guided unpooling can recover boundary details much better than other upsampling operators such as bilinear interpolation. By looking at the indices as a function of the feature map, we introduce the concept of learning to index, and present a novel index-guided encoder-decoder framework where indices are self-learned adaptively from data and are used to guide the pooling and upsampling operators, without the need of supervision. At the core of this framework is a flexible network module, termed IndexNet, which dynamically predicts indices given an input. Due to its flexibility, IndexNet can be used as a plug-in applying to any off-the-shelf convolutional networks that have coupled downsampling and upsampling stages.

	We demonstrate the effectiveness of IndexNet on the task of natural image matting where the quality of learned indices can be visually observed from predicted alpha mattes. Results on the Composition-1k matting dataset show that our model built on MobileNetv2 exhibits at least $16.1\%$ improvement over the seminal VGG-16 based deep matting baseline, with less training data and lower model capacity. Code and models has been made available at: {\small\url{https://tinyurl.com/IndexNetV1}}.

\end{abstract}

\section{Introduction}

Upsampling is an essential stage for most dense prediction tasks using deep convolutional neural networks (CNNs). The frequently used upsampling operators include transposed convolution~\cite{zeiler2014visualizing,long2015fully}, unpooling~\cite{badrinarayanan2017segnet}, \textit{periodic shuffling}~\cite{shi2016real} (also known as depth-to-space), and naive interpolation~\cite{lin2017refine,chen18v3} followed by convolution. These operators, however, are not general-purpose designs and often have different behaviors in different tasks.

\begin{figure}[t]
	\captionsetup{font=small,singlelinecheck=true}
	\begin{center}
		\includegraphics[width=\linewidth]{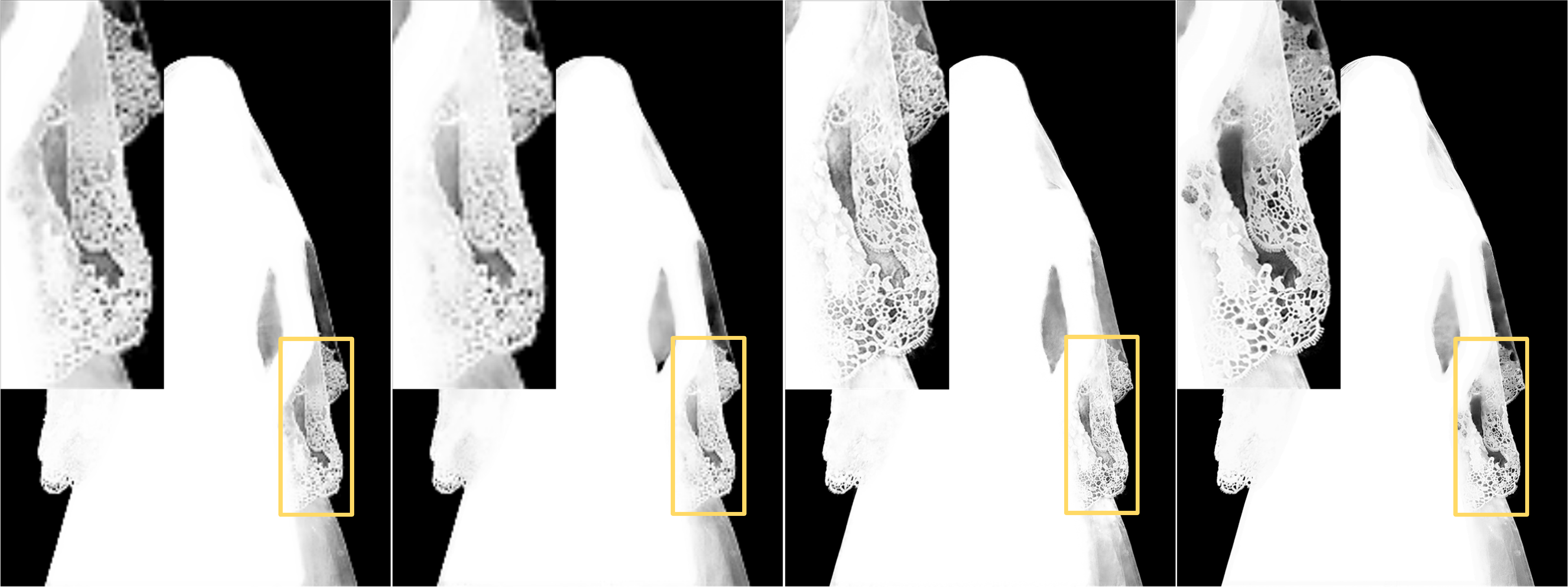}
	\end{center}
%	\vspace{-15pt}
	\caption{Alpha mattes of different models. From left to right, Deeplabv3+~\cite{chen18v3}, RefineNet~\cite{lin2017refine}, Deep Matting~\cite{xu2017deep} and Ours. %
	Bilinear upsampling fails to recover subtle details, but unpooling and our learned upsampling operator can produce much clear mattes with good local contrast.}
	\label{fig:alpha_mattes}
%	\vspace{-10pt}
\end{figure}

The widely-adopted operator in semantic segmentation or depth estimation is bilinear interpolation, rather than unpooling. A reason is that the feature map generated by unpooling is too sparse, while bilinear interpolation is likely to generate the feature map that depicts semantically-consistent regions. This is particularly true for semantic segmentation and depth estimation where pixels in a region often share the same class label or have similar depth. However, bilinear interpolation performs much worse than unpooling in boundary-sensitive tasks such as image matting. A fact is that the leading deep image matting model~\cite{xu2017deep} largely borrows the design from the SegNet~\cite{badrinarayanan2017segnet}, where unpooling is introduced. When adapting other state-of-the-art segmentation models, such as DeepLabv3+~\cite{chen18v3} and RefineNet~\cite{lin2017refine}, to this task, unfortunately, we observe both DeepLabv3+ and RefineNet fail to recover boundary details (Fig.~\ref{fig:alpha_mattes}), compared to SegNet. This makes us to ponder over \textit{what is missing} in these encoder-decoder models. After making a thorough comparison between different architectures and conducting ablative studies (Section~\ref{subsec:ablation_study}), the answer is finally made clear---\textit{indices matter}.

Compared to the bilinearly upsampled feature map, unpooling uses max-pooling indices to guide upsampling. Since boundaries in the shallow layers usually have the maximum responses, indices extracted from these responses record the boundary locations. The feature map projected by the indices thus shows improved boundary delineation. Above analyses reveal a fact that, different upsampling operators have different characteristics, and we expect a specific behavior of the upsampling operator when dealing with specific image content in a certain visual task.

It would be interesting to pose the question: \textit{Can we design a generic operator to upsample feature maps that better predict boundaries and regions simultaneously?} A key observation of this work is that max unpooling, bilinear interpolation or other \textit{upsampling operators are some forms of index functions}. For example, the nearest neighbor interpolation of a point is equivalent to allocating indices of one to its neighbor and then map the value of the point. In this sense, indices are models~\cite{kraska2018case}, therefore indices can be modeled and learned. In this work, we model indices as a function of the local feature map and learn an index function to perform upsampling within deep CNNs. In particular, we present a novel index-guided encoder-decoder framework, which naturally generalizes SegNet. Instead of using  max-pooling and unpooling, we introduce \mbox{indexed pooling} and indexed upsampling operators where downsampling and upsampling are guided by learned indices. The indices are generated dynamically conditioned on the feature map and are learned using a fully convolutional network, termed IndexNet, without supervision. IndexNet is a highly flexible module, which can be used as a plug-in applying to any off-the-shelf convolutional networks that have coupled downsampling and upsampling stages. Compared to the fixed $\max$ function, learned index functions show potentials for simultaneous boundary and region delineation.

We demonstrate the effectiveness of IndexNet on natural image matting as well as other visual tasks. In image matting, the quality of learned indices can be visually observed from predicted alpha mattes. By visualizing learned indices, we show that the indices automatically learn to capture the boundaries and textural patterns.
We further investigate alternative ways to design IndexNet, and show through extensive experiments that IndexNet can effectively improve deep image matting  both qualitatively and quantitatively. In particular, we observe that our best MobileNetv2-based~\cite{sandler2018mobilenetv2} model exhibits at least $16.1\%$ improvement against the previous best deep model, i.e., the VGG-16-based model in \cite{xu2017deep}, on the Composition-1k matting dataset.  We achieve this with using less training data, and a much more compact model, therefore significantly faster inference speed.

\section{Related Work}

We review existing widely-used upsampling operators and the main application of IndexNet---deep image matting.

%\vspace{-10pt}
\paragraph{Upsampling in Deep Networks}
Upsampling is an essential stage for almost all dense prediction tasks.
It has been intensively studied about \textit{what is the principal way to recover the resolution of the downsampled feature map} (decoding). The \textit{deconvolution} operator, also known as transposed convolution, was initially used  in~\cite{zeiler2014visualizing} to visualize convolutional activations and latter introduced to semantic segmentation \cite{long2015fully}.
To avoid checkerboard artifacts, a follow-up suggestion is the ``resize+convolution'' paradigm, which has currently become the standard configuration in state-of-the-art semantic segmentation models~\cite{chen18v3,lin2017refine}. Aside from these, \textit{perforate}~\cite{osendorfer2014image} and \textit{unpooling}~\cite{badrinarayanan2017segnet} are also two operators that generate sparse indices to guide upsampling. The indices are able to capture and keep boundary information, but the problem is that two operators induce sparsity after upsampling. Convolutional layers with large filter sizes must follow for densification. In addition, \textit{periodic shuffling} ($\mathcal{PS}$) was introduced in~\cite{shi2016real} as a fast and memory-efficient upsampling  operator for image super-resolution. $\mathcal{PS}$ recovers resolution by rearranging the feature map of size $H\times W\times Cr^2$ to $rH\times rW\times C$.

Our work is primarily inspired by the unpooling operator~\cite{badrinarayanan2017segnet}. We remark that, it is important to keep  the spatial information before loss of such information occurred in feature map downsampling, and more importantly, to use stored information during upsampling. Unpooling shows a simple and effective case of doing this, but we argue there is much room to improve. In this paper, we illustrate that the unpooling operator is a special form of index function, and we can learn an index function beyond unpooling.

\begin{figure*}[!tb]
	\captionsetup{font=small,singlelinecheck=true}
	\setlength{\abovecaptionskip}{10pt}
	\centering
	\includegraphics[width=.879\textwidth,angle=0]{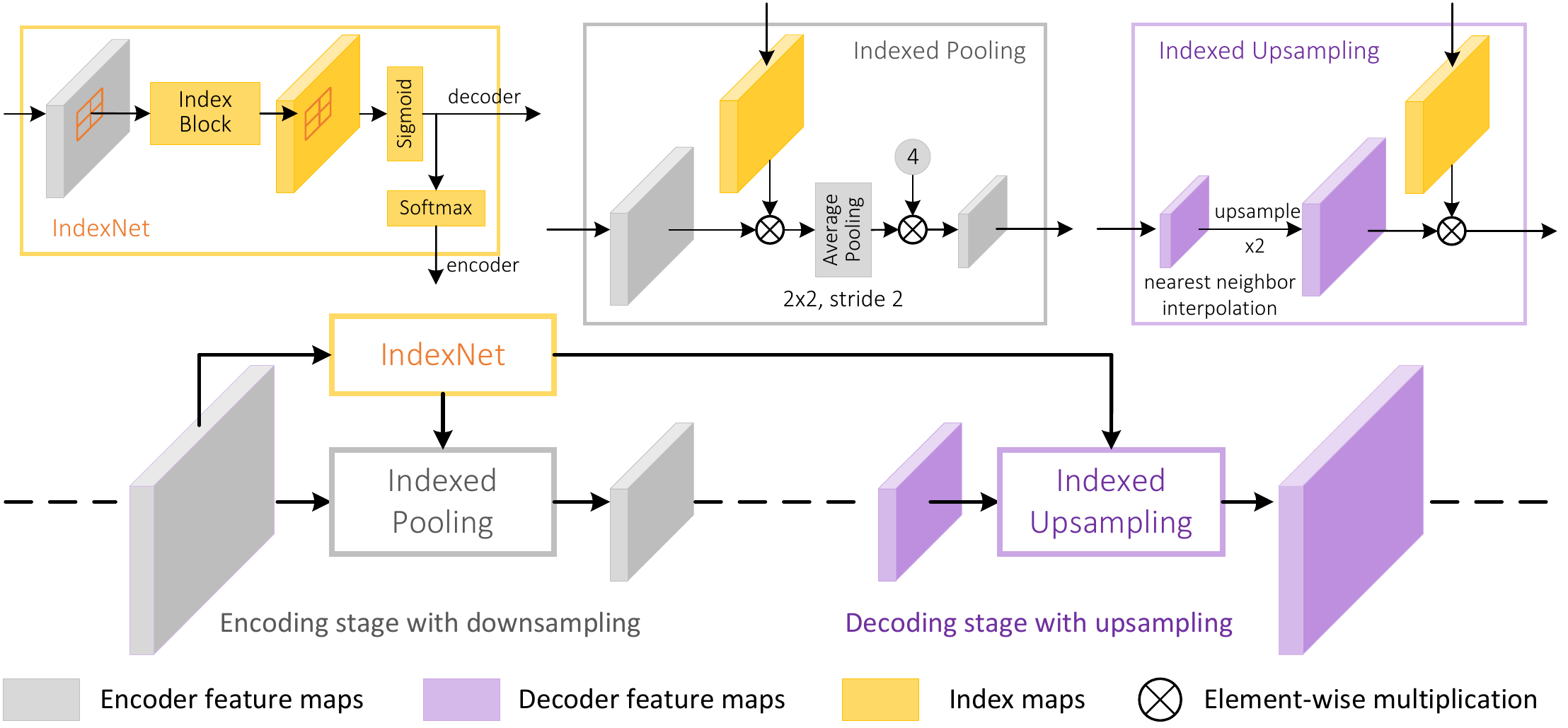}
	%\vspace{-5pt}
	\caption{Index-guided encoder-decoder framework. The proposed \mbox{IndexNet} dynamically predicts indices for individual local regions, conditional on the input local feature map itself. The predicted indices are further utilized to guide the downsampling in the encoding stage and the upsampling in corresponding decoding stage.}
	\label{fig:framework}
%	\vspace{-10pt}
\end{figure*}

%\vspace{-10pt}
\paragraph{Deep Image Matting}

In the past decades, image matting methods have been extensively studied from a low-level view~\cite{aksoy2017designing,chen2013knn,chen2013image,chuang2001bayesian,guan2006easy,he2011global,lee2011nonlocal,levin2008closed,sun2004poisson}; and particularly, they have been designed to solve the matting equation. Despite being theoretically elegant, these methods heavily rely on the color cues, rendering failures of matting in general natural scenes where colors cannot be used as reliable cues.

With the tremendous success of deep CNNs in high-level vision tasks~\cite{girshick2014rich,krizhevsky2012imagenet,long2015fully}, deep matting methods are emerging. Some initial attempts appeared in~\cite{cho2016natural} and~\cite{shen2016deep}, where classic matting approaches, such as closed-form matting~\cite{levin2008closed} and KNN matting~\cite{chen2013knn}, are still used as the backends in deep networks. Although the networks are trained end-to-end and can extract powerful features, the final performance is limited by the conventional backends. These attempts may be thought as semi-deep matting. Recently fully-deep image matting was proposed~\cite{xu2017deep}. In \cite{xu2017deep} the authors presented the first deep image matting approach based on SegNet~\cite{badrinarayanan2017segnet} and significantly outperformed other competitors. Interestingly, this SegNet-based architecture becomes the standard configuration in many recent deep matting methods~\cite{chen2018tom,chen2018human,wang2018propagation}.

SegNet is effective in matting but also computation-expensive and memory-inefficient. For instance, the inference can only be executed on CPU when testing high-resolution images, which is practically unattractive.
We show that, with our proposed IndexNet, even a lightweight backbone such as MobileNetv2-based model can surpass the VGG-16 based method in \cite{xu2017deep}.

\section{An Indexing Perspective of Upsampling}

With the argument that upsampling operators are index functions, here we offer an unified index perspective of upsampling operators. The unpooling operator is straightforward. We can define its index function in a $k\times k$ local region as an indicator function
\begin{equation}
I_{max}(x)=\mathbbm{1}(x=\max (\mat X))\,,x\in\mat X \,,
\end{equation}
where $\mat X\in\mathbb{R}^{k\times k}$. Similarly, if one extracts indices from the average pooling operator, the index function takes the form
\begin{equation}
I_{avg}(x)=\mathbbm{1}(x\in\mat X)\,.
\end{equation}
If further using $I_{avg}(x)$ during upsampling, it is equivalent to the nearest neighbor interpolation. Regarding the bilinear interpolation and deconvolution operators, their index functions have an identical form
\begin{equation}
I_{bilinear/dconv}(x)=\mat W\otimes\mathbbm{1}(x\in\mat X)\,,
\end{equation}
where $\mat W$ is the weight/filter of the same size as $\mat X$, and $\otimes$ denotes the element-wise multiplication. The difference is that, $\mat W$ in deconvolution is learned, while $\mat W$ in bilinear interpolation stays fixed. Indeed, bilinear upsampling has been shown to be a special case of deconvolution~\cite{long2015fully}. Notice that, in this case, the index function generates soft indices. The sense of index for the $\mathcal{PS}$ operator~\cite{shi2016real} is even much clear, because the rearrangement of the feature map per se is an indexing process. Considering $\mathcal{PS}$ a tensor $\mathcal{Z}$ of size $1\times1\times r^2$ to a matrix $\mat Z$ of size $r\times r$, the index function can be expressed by the one-hot encoding
\begin{equation}
I^l_{ps}(x)=\mathbbm{1}(x= \mathcal{Z}_l)\,,l=1,...,r^2\,,
\end{equation}
such that $\mat Z_{m,n}=\mathcal{Z}[I^l_{ps}(x)]$, where $m=1,...,r$, $n=1,...,r$, and $l=(r-1)*m+n$. $\mathcal{Z}_l$ denotes the $l$-th element of $\mathcal{Z}$. A similar notation applies to $\mat Z_{m,n}$.

Since upsampling operators can be unified by the notion of index function, in theory it is possible to learn an index function that adaptively captures local spatial patterns.

\section{\mbox{Index-Guided Encoder-Decoder Framework}}

Our framework is a natural generalization of SegNet, as schematically illustrated in Fig.~\ref{fig:framework}. For ease of exposition, we assume the downsampling and upsampling rates are $2$, and the pooling operator has a kernel size of $2\times2$. At the core of our framework is the IndexNet module that dynamically generates indices given the feature map. The proposed indexed pooling and indexed upsampling operators further receive generated indices to guide the downsampling and upsampling, respectively. In practice, multiple such modules can be combined and used analogues to the max pooling layers. We provide details as follows.

\subsection{Learning to Index, to Pool, and to Upsample}

\noindent\textbf{IndexNet} models the index as a function of the feature map $\mathcal{X}\in\mathbb{R}^{H\times W\times C}$. It generates two index maps for downsampling and upsampling given the input $\mathcal{X}$. An important concept for the index is that an index can either be represented in a natural order, e.g., 1, 2, 3, ..., or be represented in a logical form, i.e., 0, 1, 0, ..., which means an index map can be used as a mask. In fact, this is how we use the index map in downsampling and upsampling. The predicted index shares the same physical notation of the index in computer science, except that we generate \textit{soft} indices for smooth optimization, i.e., for any index $i$, $i\in[0,1]$.

\begin{figure}[!tb]
	\captionsetup{font=small,singlelinecheck=true}
	\setlength{\abovecaptionskip}{10pt}
	\centering
	\includegraphics[width=3in,angle=0]{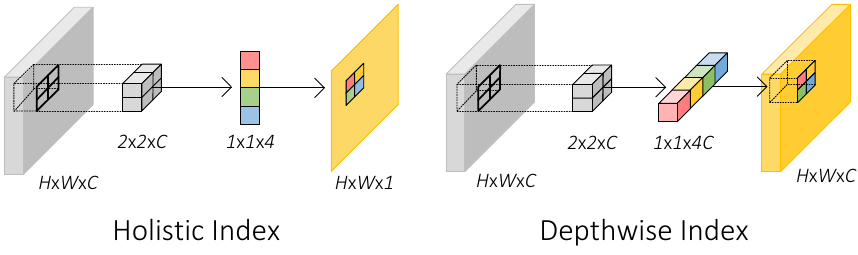}
	%\vspace{-10pt}
	\caption{Conceptual differences between holistic index and depthwise index.}
	\label{fig:holistic_depthwise}
%	\vspace{-15pt}
\end{figure}

IndexNet consists of a predefined index block and two index normalization layers. An index block can  simply be a heuristically defined function, e.g., a $\max$  function, or more generally, a neural network. In this work, the index block is designed to use a fully convolutional network. According to the shape of the output index map, we investigate two families of index networks: \textit{holistic index networks (HINs)} and \textit{depthwise (separable) index networks (DINs)}. Their conceptual differences are shown in Fig.~\ref{fig:holistic_depthwise}. HINs learn an index function $I(\mathcal{X}):\mathbb{R}^{H\times W\times C}\rightarrow\mathbb{R}^{H\times W\times1}$. In this case, all channels of the feature map share a holistic index map. In contrast, DINs learn an index function $I(\mathcal{X}):\mathbb{R}^{H\times W\times C}\rightarrow\mathbb{R}^{H\times W \times C}$, where the index map is of the same size as the feature map. We will discuss concrete design of index networks in Sections~\ref{subsec:holistic_networks} and~\ref{subsec:depthwise_networks}.

Note that the index map sent to the encoder and decoder are normalized differently. The decoder index map only goes through a \textit{sigmoid} function such that for any predicted index $i\in(0,1)$. As for the encoder index map, indices of a local region $L$ are further normalized by a \textit{softmax} function such that $\sum_{i\in{L}}i=1$. The reason behind the second normalization is to guarantee the magnitude consistency of the feature map after downsampling.

\vspace{5pt}
\noindent\textbf{Indexed Pooling} ($\mathcal{IP}$) executes downsampling using generated indices. Given a local region $E\in\mathbb{R}^{k\times k}$, $\mathcal{IP}$ calculates a weighted sum of activations and corresponding indices over $E$ as $\mathcal{IP}(E)=\sum_{x\in E}I(x)x$, where $I(x)$ is the index of $x$. It is easy to infer that max pooling and average pooling are both special cases of $\mathcal{IP}$. In practice, this operator can be easily implemented with an element-wise multiplication between the feature map and the index map, an average pooling layer, and a multiplication of a constant, as instantiated in Fig.~\ref{fig:framework}.

\vspace{5pt}
\noindent\textbf{Indexed Upsampling} ($\mathcal{IU}$) is the inverse operator of $\mathcal{IP}$. $\mathcal{IU}$ upsamples $d\in\mathbb{R}^{1\times 1}$ that spatially corresponds to $E$ taking the same indices into account. Let $I\in\mathbb{R}^{k\times k}$ be the local index map formed by $I(x)$s, $\mathcal{IU}$ upsamples $d$ as $\mathcal{IU}(d)=I\otimes D$, where $\otimes$ denotes the element-wise multiplication, and $D$ is of the same size as $I$ and is upsampled from $d$ with the nearest neighbor interpolation. An important difference between deconvolution and $\mathcal{IU}$ is that, deconvolution applies a fixed kernel to all local regions, even if the kernel is learned, while $\mathcal{IU}$ upsamples different regions with different kernels (indices).

\subsection{Holistic Index Networks}\label{subsec:holistic_networks}

Here we instantiate two types of HINs. Recall that HINs learn an index function $I(\mathcal{X}):\mathbb{R}^{H\times W\times C}\rightarrow\mathbb{R}^{H\times W\times1}$. A naive design choice is to assume a linear relationship between the feature map and the index map.

\begin{figure}[!tb]
	\captionsetup{font=small,singlelinecheck=true}
	\setlength{\abovecaptionskip}{10pt}
	\centering
	\includegraphics[width=2.4in,angle=0]{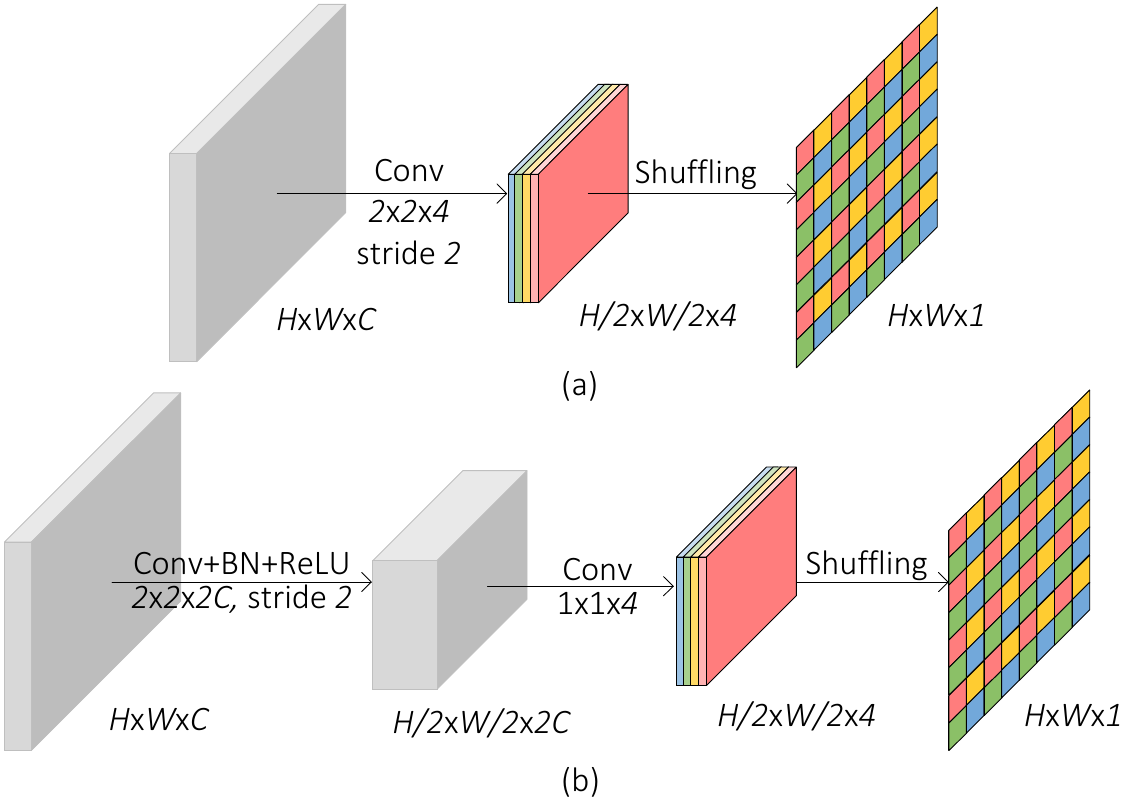}\vspace{-10pt}
	\caption{Holistic index networks. (a) a linear index network; (b) a nonlinear index network.}
	\label{fig:holistic_networks}
%	\vspace{-15pt}
\end{figure}

\vspace{5pt}
\noindent\textit{Linear Holistic Index Networks.} An example is shown in Fig.~\ref{fig:holistic_networks}(a). The network is implemented in a fully convolutional way. It first applies $2$-stride $2\times2$ convolution to the feature map of size $H\times W\times C$, generating a concatenated index map of size $H/2\times W/2\times 4$. Each slice of the index map ($H/2\times W/2\times 1$) is designed to correspond to the indices of a certain position of all local regions, e.g., the top-left corner of all $2\times 2$ regions. The network finally applies a $\mathcal{PS}$-like shuffling operator to rearrange the index map to the size of $H\times W\times 1$.

In many situations, assuming a linear relationship is not sufficient. An obvious fact is that a linear function even cannot fit the $\max$ function. Naturally the second design choice is to add nonlinearity into the network.

\vspace{5pt}
\noindent\textit{Nonlinear Holistic Index Networks.} Fig.~\ref{fig:holistic_networks}(b) illustrates a nonlinear HIN where the feature map is first projected to a map of size $H/2\times W/2\times 2C$, followed by a batch normalization layer and a ReLU function for nonlinear mappings. We then use point-wise convolution to reduce the channel dimension to an indices-compatible size. The rest transformations follow its linear counterpart.

\vspace{5pt}
\noindent\textbf{Remark 1}. Note that, the holistic index map is shared by all channels of the feature map, which means the index map should be expanded to the size of $H\times W\times C$ when feeding into $\mathcal{IP}$ and $\mathcal{IU}$. Fortunately, many existing packages support implicit expansion over the singleton dimension. This index map could be thought as a collection of local attention maps~\cite{mnih2014recurrent} applied to individual local spatial regions. In this case, the $\mathcal{IP}$ and $\mathcal{IU}$ operators can also be referred to ``attentional pooling'' and ``attentional upsampling''.

\begin{figure}[!tb]
	\captionsetup{font=small,singlelinecheck=true}
	\setlength{\abovecaptionskip}{10pt}
	\centering
	\includegraphics[width=3.2in,angle=0]{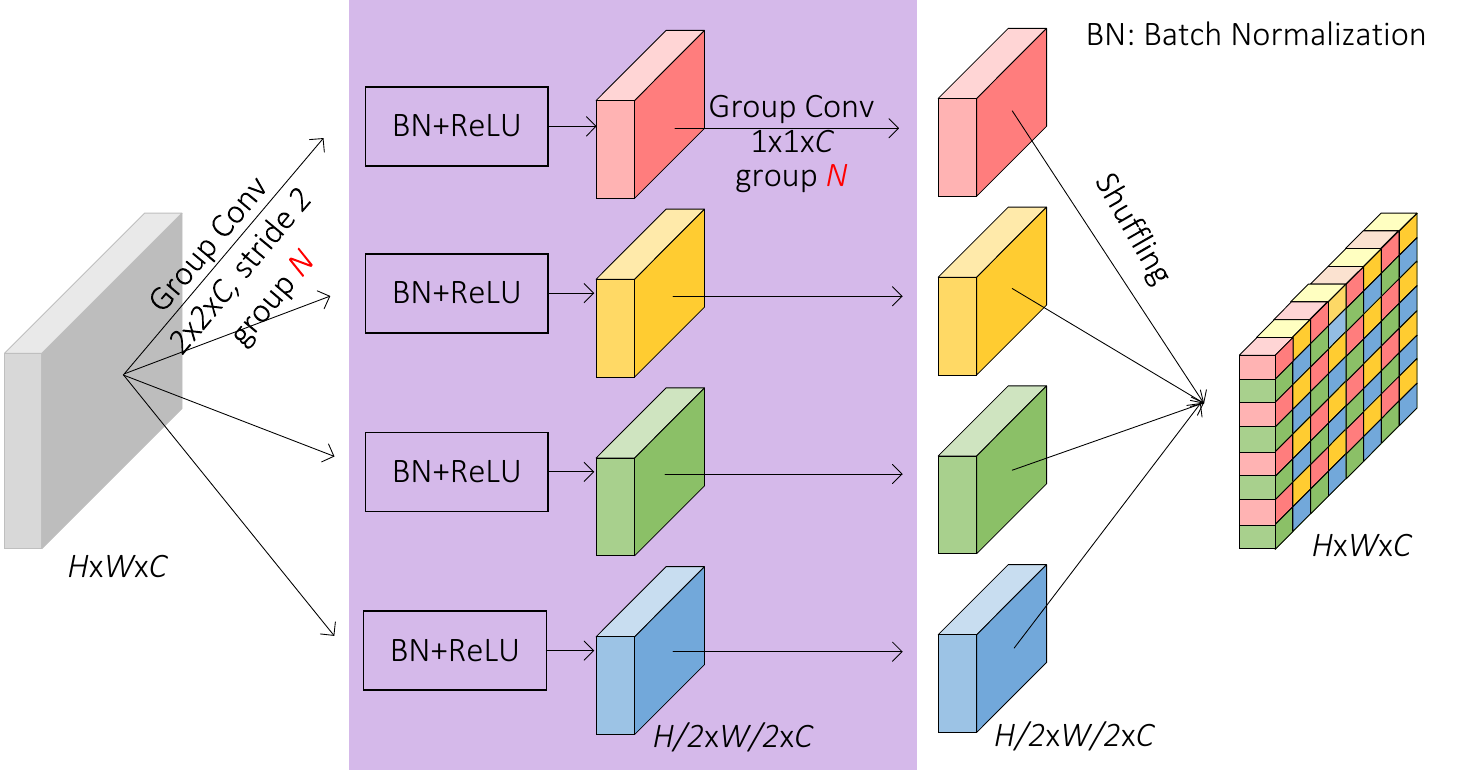}\vspace{-8pt}
	\caption{Depthwise index networks. $N=C$ for the O2O assumption, and $N=1$ for the M2O. The masked modules are invisible to linear networks.}
	\label{fig:depthwise_networks}
%	\vspace{-15pt}
\end{figure}

\subsection{Depthwise Index Networks}\label{subsec:depthwise_networks}

In DINs, we find $I(\mathcal{X}):\mathbb{R}^{H\times W\times C}\rightarrow\mathbb{R}^{H\times W \times C}$, i.e., each spatial index corresponds to each spatial activation. This family of networks further has two high-level design strategies that correspond to two different assumptions.

\vspace{3pt}
\noindent\textbf{One-to-One (O2O) Assumption} assumes that each slice of the index map only relates to its corresponding slice of the feature map. It can be denoted by a local index function $l(\mathcal{X}): \mathbb{R}^{k\times k\times1}\rightarrow\mathbb{R}^{k\times k\times1}$, where $k$ denotes the size of local region. Similar to HINs, DINs can also be designed to have linear/nonlinear modeling ability. Fig.~\ref{fig:depthwise_networks} shows an example when $k=2$. Note that, different from HINs, DINs follow a multi-column architecture. Each column predicts indices specific to a certain spatial location of all local regions. The O2O assumption can be easily satisfied in DINs with grouped convolution.

\vspace{3pt}
\noindent\textit{Linear Depthwise Index Networks.}  As per Fig.~\ref{fig:depthwise_networks}, a feature map goes through four parallel convolutional layers with the same kernel size of $2\times2\times C$, a stride of $2$, and $C$ groups, leading to four downsampled feature maps of size $H/2\times W/2 \times C$. The final index map is composed from the four feature maps by shuffling and rearrangement. Note that the parameters of four convolutional layers are not shared.

\vspace{3pt}
\noindent\textit{Nonlinear Depthwise Index Networks.} Nonlinear DINs can be easily modified from linear DINs by inserting four extra convolutional layers. Each of them is followed by a BN layer and a ReLU unit, as shown in Fig.~\ref{fig:depthwise_networks}. The rest remains the same as the linear DINs.

\vspace{3pt}
\noindent\textbf{Many-to-One (M2O) Assumption} assumes that each slice of the index map relates with all channels of the feature map. The local index function is defined as $l(\mathcal{X}): \mathbb{R}^{k\times k\times C}\rightarrow\mathbb{R}^{k\times k\times1}$. Compared to O2O DINs, the only difference in implementation is the use of standard convolution instead of group convolution, i.e., $N=1$ in Fig.~\ref{fig:depthwise_networks}.

\vspace{3pt}
\noindent\textbf{Learning with Weak Context}. A desirable property of IndexNet is that it can predict indices even from a large local feature map, e.g., $l(\mathcal{X}): \mathbb{R}^{2k\times 2k\times C}\rightarrow\mathbb{R}^{k\times k\times1}$. An intuition behind this idea is that, if one identifies a local maximum point from a $k\times k$ region, its surrounding $2k\times2k$ region can further support whether this point is a part of a boundary or just an isolated noise point. This idea can be easily implemented by enlarging the convolutional kernel and is also applicable to HINs.

\vspace{3pt}
\noindent\textbf{Remark 2}. Both HINs and DINs have merits and drawbacks. It is clear that DINs have higher capacity than HINs, so DINs may capture more complex local patterns but also be at a risk of overfitting. By contrast, the index map generated by HINs is shared by all channels of the feature map, so the decoder feature map can reserve its expressibility without forcibly reducing its dimensionality to fit the shape of the index map during upsampling. This gives much flexibility for decoder design, while it is not the case for DINs.

\subsection{Relation to Other Networks}

If considering the dynamic property of IndexNet, \mbox{IndexNet} shares a similar spirit with some recent networks.

\vspace{3pt}
\noindent\textbf{Spatial Transformer Networks (STNs)}~\cite{jaderberg2015spatial}. The STN learns dynamic spatial transformation by regressing desired transformation parameters $\theta$ with a localized network. A spatially-transformed output is then produced by a sampler parameterized by $\theta$. Such a transformation is holistic for the feature map, which is similar to HINs. The differences between STN and IndexNet are that their learning targets have different physical definitions (spatial transformations vs. spatial indices), and that, STN is designed for global transformation, while IndexNet predicts local indices.

\vspace{3pt}
\noindent\textbf{Dynamic Filter Networks (DFNs)}~\cite{jia2016dynamic}. The DFN dynamically generates filter parameters on-the-fly with a so-called filter generating network. Compared to conventional filter parameters that are initialized, learned, and stayed fixed during inference, filter parameters in DFN are dynamic and sample-specific. The main difference between DFN and IndexNet lies in the motivation of the design. Dynamic filters are learned for adaptive feature extraction, but learned indices are used for dynamic downsampling and upsampling.

\vspace{3pt}
\noindent\textbf{Deformable Convolutional Networks (DCNs)}~\cite{dai2017deformable}. The DCN introduces deformable convolution and deformable RoI pooling. The key idea is to predict offsets for convolutional and pooling kernels, so DCN is also a dynamic network. While these convolution and pooling operators concern spatial transformations, they are still built upon standard max pooling and are not designed for upsampling purposes. By contrast, index-guided $\mathcal{IP}$ and $\mathcal{IU}$ are fundamental operators and may be integrated into RoI pooling.

\vspace{3pt}
\noindent\textbf{Attention Networks}~\cite{mnih2014recurrent}. Attention networks are a broad family of networks that adopt attention mechanisms. The mechanisms introduce multiplicative interactions between inferred attention maps and feature maps. In Computer Vision, these mechanisms often refer to spatial attention~\cite{wang2017residual}, channel attention~\cite{hu2018squeeze} or both~\cite{woo2018cbam}. As aforementioned, $\mathcal{IP}$ and $\mathcal{IU}$ in HINs can be viewed as attentional operators to some extent, which means indices are attention. In a reverse sense, attention is also indices. For example, max-pooling indices are a form of hard attention. Indices offer a new perspective to understand attention. It is worth noting that, despite IndexNet in its current implementation closely relates to attention, it has a distinct physical definition and specializes in upsampling rather than refining feature maps.

\section{Results and Discussions}

We evaluate our framework and IndexNet on the task of image matting. This task is particularly suitable for visualizing the quality of learned indices. We mainly conduct experiments on the Adobe Image Matting dataset~\cite{xu2017deep}. This is so far the largest publicly available matting dataset. The training set has 431 foreground objects and ground-truth alpha mattes.\footnote{The original paper reported that there were 491 images, but the released dataset only includes 431 images.
As a result, we
use fewer
training data
than the original paper.}
Each foreground is composited with 100 background images randomly chosen from MS COCO~\cite{lin2014microsoft}. The test set termed Composition-1k includes 100 unique objects. Each of them is composited with 10 background images chosen from Pascal VOC~\cite{everingham2010pascal}. Overall, we have 43100 training images and 1000 testing images. We evaluate the results using widely-used Sum of Absolute Differences (SAD), Mean Squared Error (MSE), and perceptually-motivated Gradient (Grad) and Connectivity (Conn) errors~\cite{rhemann2009perceptually}. The evaluation code implemented by~\cite{xu2017deep} is used. In what follows, we first describe our modified MobileNetv2-based architecture and training details. We then perform extensive ablation studies to justify choices of model design, make comparisons of different index networks, and visualize learned indices. We also report performance on the $\tt alphamatting.com$ online benchmark~\cite{rhemann2009perceptually} and extend IndexNet to other visual tasks.

\subsection{Implementation Details}

Our implementation is based on PyTorch~\cite{paszke2017automatic}. Here we describe the network architecture used and some essential training details.
\begin{figure}[!tb]
	\captionsetup{font=small,singlelinecheck=true}
	\setlength{\abovecaptionskip}{10pt}
	\centering
	\includegraphics[width=2.8in,angle=0]{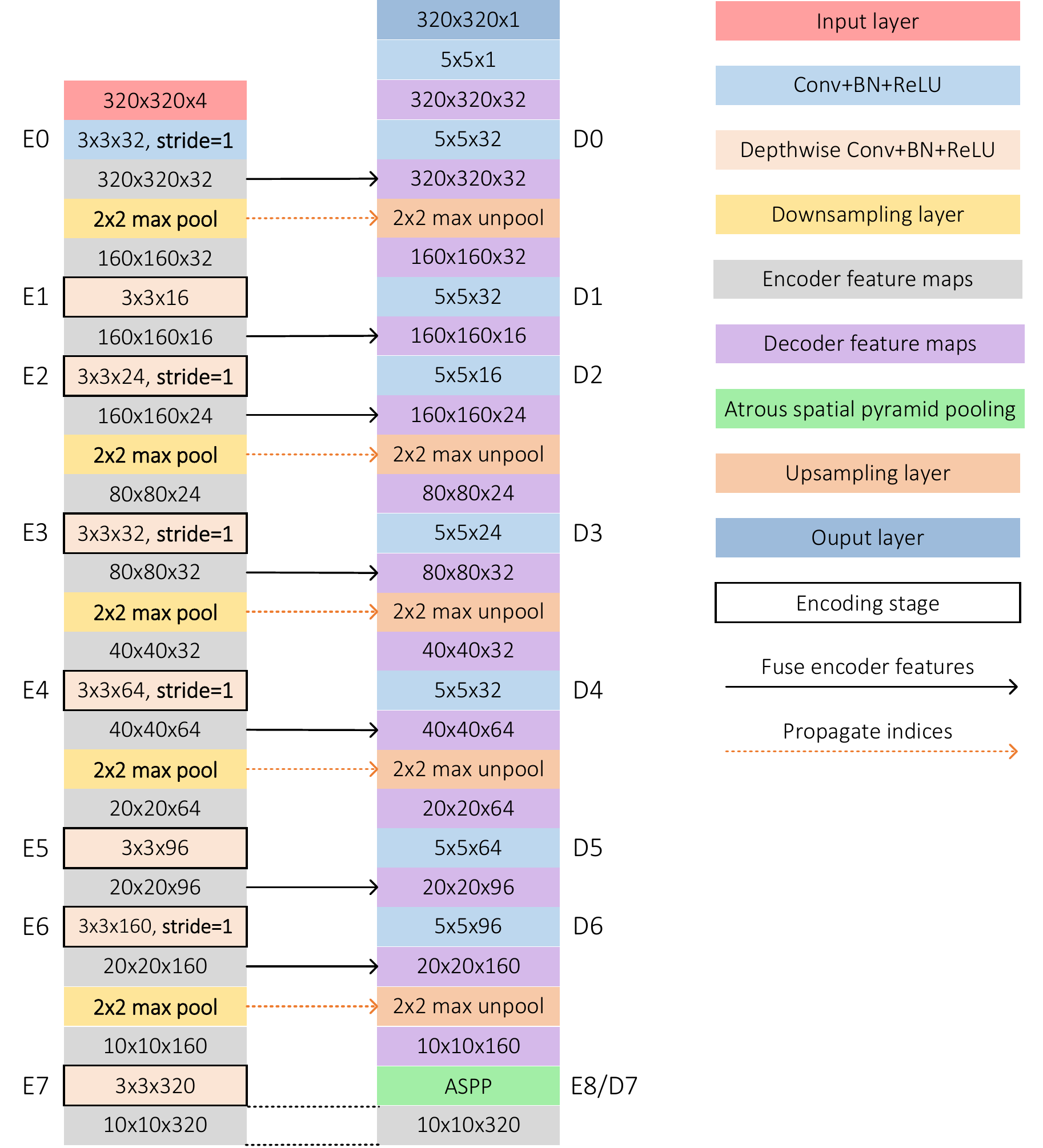}\vspace{-5pt}
	\caption{Customized MobileNetv2-based encoder-decoder network architecture. Our modifications are boldfaced.}
	\label{fig:network_architecture}
%	\vspace{-10pt}
\end{figure}
\begin{table*}[!t] \small
	\captionsetup{font=small,singlelinecheck=true}
	\centering
	\addtolength{\tabcolsep}{3pt}
	\renewcommand\arraystretch{1.0}
	\begin{tabular}{llccccc|ccccc}
			\hline
			No. & Architecture					& Backbone		& Fusion 	& Indices 	& Context 	& OS 	& SAD 	& MSE 	& Grad & Conn \\
			\hline
			B1	& DeepLabv3+~\cite{chen18v3}	& MobileNetv2	& Concat	& No		& ASPP		& 16	& 60.0	& 0.020	& 39.9 & 61.3 \\
			B2 	& RefineNet~\cite{lin2017refine}& MobileNetv2	& Skip 		& No 		& CRP 		& 32 	& 60.2	& 0.020	& 41.6 & 61.4 \\
			\rowcolor{mygray}
			B3 	& SegNet~\cite{xu2017deep}		& VGG16			& No		& Yes		& No		& 32	& \textbf{54.6} 	& \textbf{0.017} & 36.7 & 55.3 \\
			\rowcolor{mygray}
			B4 	& SegNet						& VGG16			& No		& No		& No		& 32	& 122.4 & 0.100 & 161.2 & 130.1 \\
			\rowcolor{mygray}
			B5 	& SegNet		 				& MobileNetv2	& No 		& Yes 		& No 		& 32	& 60.7	& 0.021	& 40.0 & 61.9 \\
			\rowcolor{mygray}
			B6 	& SegNet		 				& MobileNetv2	& No 		& No 		& No 		& 32	& 78.6  & 0.031 & 101.6& 82.5 \\
			B7 	& SegNet		 				& MobileNetv2	& No 		& Yes 		& ASPP 		& 32	& 58.0	& 0.021	& 39.0 & 59.5 \\
			B8 	& SegNet		 				& MobileNetv2	& Skip 		& Yes 		& No 		& 32	& 57.1 	& 0.019	& 36.7 & 57.0 \\
			B9 	& SegNet		 				& MobileNetv2	& Skip 		& Yes 		& ASPP 		& 32	& 56.0	& \textbf{0.017}	& 38.9 & 55.9 \\
			B10 & UNet	 						& MobileNetv2	& Concat 	& Yes 		& No 		& 32 	& 54.7	& \textbf{0.017}	& 34.3 & \textbf{54.7} \\
			B11 & UNet	 						& MobileNetv2	& Concat 	& Yes 		& ASPP 		& 32	& 54.9	& \textbf{0.017} & \textbf{33.8} & 55.2 \\
			\hline
		\end{tabular}
	\vspace{-8pt}
	\caption{Ablation study of design choices. Fusion: fuse encoder features; Indices: max-pooling indices (when Indices is `No', bilinear interpolation is used for upsampling); CRP: chained residual pooling~\cite{lin2017refine}; ASPP: atrous spatial pyramid pooling~\cite{chen18v3}; OS: output stride.
	The lowest errors are boldfaced.}
	\label{tab:architecture}
	%
%	\vspace{-12pt}
\end{table*}

\vspace{3pt}
\noindent\textbf{Network Architecture}. We build our model based on MobileNetv2~\cite{sandler2018mobilenetv2} with only slight modifications to the backbone. An important reason why we choose MobileNetv2 is that this lightweight model allows us to infer high-resolution images on a GPU, while other high-capacity backbones cannot. The basic network configuration is shown in Fig.~\ref{fig:network_architecture}. It also follows the encoder-decoder paradigm same as SegNet. We simply change all 2-stride convolution to be 1-stride and attach 2-stride $2\times2$ max pooling after each encoding stage for downsampling, which allows us to extract indices. If applying the IndexNet idea, max pooling and unpooling layers can be replaced with $\mathcal{IP}$ and $\mathcal{IU}$, respectively. We also investigate alternative ways for low-level feature fusion and whether encoding context (Section~\ref{subsec:ablation_study}). Notice that, the matting refinement stage~\cite{xu2017deep} is not considered in this paper.

\vspace{3pt}
\noindent\textbf{Training Details}. To enable a direct comparison with deep matting~\cite{xu2017deep}, we follow the same training configurations used in~\cite{xu2017deep}. The 4-channel input concatenates the RGB image and its trimap. We follow exactly the same data augmentation strategies, including $320\times320$ random cropping, random flipping, random scaling, and random trimap dilation. All training samples are created on-the-fly. We use a combination of the alpha prediction loss and the composition loss during training as in~\cite{xu2017deep}. Only losses from the unknown region of the trimap are calculated. Encoder parameters are pretrained on ImageNet~\cite{deng2009imagenet}. Note that, the parameters of the $4$-th input channel are initialized with zeros. All other parameters are initialized with the improved Xavier~\cite{he2015delving}. The Adam optimizer~\cite{kingma2015adam} is used. We update parameters with $30$ epochs (around $90,000$ iterations). The learning rate is initially set to $0.01$ and reduced by $10\times$ at the $20$-th and $26$-th epoch respectively. We use a batch size of $16$ and fix the BN layers of the backbone.

\subsection{Adobe Image Matting Dataset}
\label{subsec:ablation_study}

\vspace{3pt}
\noindent\textbf{Ablation Study on Model Design}. Here we investigate strategies for fusing low-level features (no fusion, skip fusion as in ResNet~\cite{he2016deep} or concatenation as in UNet~\cite{ronneberger2015u}) and whether encoding context for image matting. $11$ baselines are consequently built to justify model design. Results on the Composition-1k testing set are reported in Table~\ref{tab:architecture}. B3 is cited from~\cite{xu2017deep}. We can make the following observations: \textbf{i}) Indices are of great importance. Matting can significantly benefit from only indices (B3 vs.\ B4, B5 vs.\ B6); \textbf{ii}) State-of-the-art semantic segmentation models cannot be directly applied to image matting (B1/B2 vs.\  B3); \textbf{iii}) Fusing low-level features help, and concatenation works better than the skip connection but at a cost of increased computation (B5 vs.\ B8 vs.\ B10 or B7 vs.\ B9 vs.\ B11); \textbf{iv}) Our intuition tells that the context may not help a low-level task like matting, while results show that encoding context is generally encouraged (B5 vs.\  B7 or B8 vs.\ B9 or B10 vs.\  B11). Indeed, we observe that the context sometimes can help to improve the quality of the background; \textbf{v}) A MobileNetv2-based model can work as well as a VGG-16-based one with appropriate design choices (B3 vs.\ B11).

For the following experiments, we now mainly use B11.

\begin{figure*}[!tb]
	\captionsetup{font=small,singlelinecheck=true}
	\setlength{\abovecaptionskip}{10pt}
	\centering
	\includegraphics[width=\linewidth,angle=0]{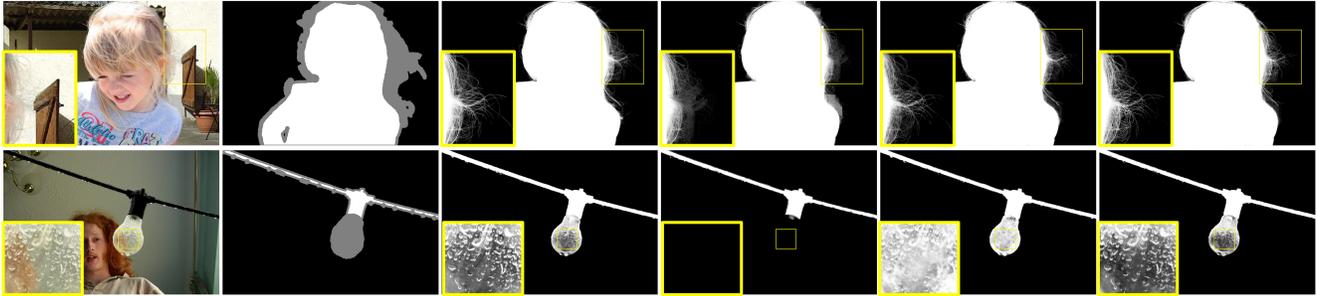}\vspace{-8pt}
	\caption{Qualitative results on the Composition-1k testing set. From left to right, the original image, trimap, ground-truth alpha matte, closed-form matting~\cite{levin2008closed}, deep image image~\cite{levin2008closed}, and ours (M2O DIN with ``nonlinear + context'').
	%See the Supplement for further results.
	}
	\label{fig:composition-1k-results}
	%\vspace{-5pt}
\end{figure*}

\vspace{3pt}
\noindent\textbf{Ablation Study on Index Networks}. Here we compare different index networks and justify their effectiveness. The configurations of index networks used in the experiments follow Figs.~\ref{fig:holistic_networks} and~\ref{fig:depthwise_networks}. We primarily investigate the $2\times2$ kernel with a stride of $2$. Whenever the weak context is considered, we use a $4\times4$ kernel in the first convolutional layer of index networks. To highlight the effectiveness of HINs, we further build a baseline called \textit{holistic max index (HMI)} where max-pooling indices are extracted from a squeezed feature map $\mathcal{X}'\in\mathbb{R}^{H\times W\times1}$. $\mathcal{X}'$ is generated by applying the max function along the channel dimension of $\mathcal{X}\in\mathbb{R}^{H\times W\times C}$. We also report the performance when setting the width multiplier of MobileNetV2 used in B11 to be $1.4$ (B11-1.4). This allows us to justify whether the improved performance is due to increased model capacity.
Results on the \mbox{Composition-1k} testing dataset are listed in Table~\ref{tab:index_function}. We observe that, except the most naive linear HIN, all index networks consistently reduce the errors. In particular, nonlinearity and the context generally have a positive effect on deep image matting. Compared to HMI, the direct baseline of HINs, the best HIN (``Nonlinear+Context'') has at least $12.3\%$ relative improvement. Compared to B11, the baseline of DINs, M2O DIN with ``Nonlinear+Context'' exhibits at least $16.5\%$ relative improvement. Notice that, our best model even outperforms the state-of-the-art DeepMatting~\cite{xu2017deep} that has the refinement stage, and is also computationally efficient with less memory consumption---the inference can be performed on the GTX 1070 over $1920\times1080$ high-resolution images. Some qualitative results are shown in Fig.~\ref{fig:composition-1k-results}. Our predicted mattes show improved delineation for edges and textures like hair and water drops.

\begin{table}[!t] \footnotesize
	\captionsetup{font=small,singlelinecheck=true}
	\centering
%	\addtolength{\tabcolsep}{-1pt}
	\renewcommand\arraystretch{1.0}
	\begin{tabular}{>{\centering}p{.3cm}|>{\centering}p{.3cm}|c|c|c|c|c|c}
			\hline
			\multicolumn{2}{c|}{Method} 				& \#Param. & GFLOPs & SAD & MSE & Grad & Conn\\
			\hline
			\multicolumn{2}{l|}{B3~\cite{xu2017deep}} 	& 130.55M & 32.34 & 54.6 & 0.017 & 36.7 & 55.3 \\
			\multicolumn{2}{l|}{B11} 					& 3.75M	  & 4.08  & 54.9 & 0.017 & 33.8 & 55.2 \\
			\multicolumn{2}{l|}{B11-1.4} 				& 8.86M	  & 7.61  & 55.6 & 0.016 & 36.4 & 55.7 \\
			\multicolumn{2}{l|}{HMI} 					& 3.75M   & 4.08  & 56.5 & 0.021 & 33.0 & 56.4 \\
			\hline
			NL & C & $\Delta$ \\
			\hline
			& & \multicolumn{6}{c}{HINs}\\
			\hline
			& 									 		& +4.99K  & 4.09  & 55.1 & 0.018 & 32.1 & 55.2\\
			& \checkmark                                & +19.97K & 4.11  & 53.5 & 0.018 & 31.0 & 53.5\\
			\checkmark & 						 		& +0.26M  & 4.22  & 50.6 & 0.015 & 27.9 & 49.4\\
			\checkmark & \checkmark 			 		& +1.04M  & 4.61  & 49.5 & 0.015 & \textbf{25.6} & 49.2\\
			\hline
			& & \multicolumn{6}{c}{O2O DINs}\\
			\hline
			&          					                & +4.99K  & 4.09  & 50.3 & 0.015 & 33.7 & 50.0\\
			& \checkmark                                & +19.97K & 4.11  & 47.8 & 0.015 & 26.9 & 45.6\\
			\checkmark &            	                & +17.47K & 4.10  & 50.6 & 0.016 & 26.5 & 50.3\\
			\checkmark & \checkmark 	                & +47.42K & 4.15  & 50.2 & 0.016 & 26.8 & 49.3\\
			\hline
			& & \multicolumn{6}{c}{M2O DINs}\\
			\hline
			&          				             		& +0.52M & 4.34   & 51.0 & 0.015 & 33.7 & 50.5\\
			& \checkmark                                & +2.07M & 5.12   & 50.6 & 0.016 & 31.9 & 50.2\\
			\checkmark &                         		& +1.30M & 4.73   & 48.9 & 0.015 & 32.1 & 47.9\\
			\checkmark & \checkmark              		& +4.40M & 6.30   & \textbf{45.8} & \textbf{0.013} & 25.9 & \textbf{43.7}\\
			\hline
			\hline
			\multicolumn{4}{l|}{Closed-Form~\cite{levin2008closed}} & 168.1 & 0.091 & 126.9 & 167.9 \\
			\multicolumn{4}{l|}{DeepMatting w. Refinement~\cite{xu2017deep}} & 50.4 & 0.014 & 31.0 & 50.8 \\
			\hline
		\end{tabular}
	%
	%\vspace{-8pt}
	\caption{Results on the Composition-1k testing set. GFLOPs are measured on a $224\times224\times4$ input. NL: Non-Linearity; C: Context. The lowest errors are boldfaced.}
	\label{tab:index_function}
	%
	%\vspace{-12pt}
\end{table}

\begin{table*}[!t] \scriptsize
	\captionsetup{font=small,singlelinecheck=false}
	\centering
	\addtolength{\tabcolsep}{-3pt}
	\renewcommand\arraystretch{1.0}
	\begin{tabular}{l|c|ccc|ccc|ccc|ccc|ccc|ccc|ccc|ccc|ccc}
		\hline
		\multirow{2}{*}{Gradient Error} & \multicolumn{4}{c|}{Average Rank}	& \multicolumn{3}{c|}{Troll}	& \multicolumn{3}{c|}{Doll} & \multicolumn{3}{c|}{Donkey} & \multicolumn{3}{c|}{Elephant}	& \multicolumn{3}{c|}{Plant} 	& \multicolumn{3}{c|}{Pineapple} 	& \multicolumn{3}{c|}{Plastic Bag} 	& \multicolumn{3}{c}{Net} \\
		& Overall & S & L & U & S & L & U & S & L & U & S & L & U & S & L & U & S & L & U & S & L & U & S & L & U & S & L & U\\
		\hline
		IndexNet Matting & \textbf{9} & \textbf{7.3} & \textbf{7.6} & \textbf{12.3} & \textbf{0.2} & \textbf{0.2} & \textbf{0.2} & \textbf{0.1} & \textbf{0.1} & 0.3 & 0.2 & 0.2 & \textbf{0.2} & \textbf{0.2} & \textbf{0.2} & \textbf{0.4} & 1.7 & 1.9 & 2.5 & 1 & 1.1 & \textbf{1.3} & 1.1 & 1.2 & 1.2 & 0.4 & 0.5 & \textbf{0.5}\\
		\rowcolor{mygray}
		AlphaGAN~\cite{lutz2018alphagan} & 13.2 & 12 & 10.8 & 16.8 & 0.2 & 0.2 & 0.2 & 0.2 & 0.2 & 0.3 & 0.2 & 0.3 & 0.3 & 0.2 & 0.2 & 0.4 & 1.8 & 2.4 & 2.7 & 1.1 & 1.4 & 1.5 & 0.9 & 1.1 & 1 & 0.5 & 0.5 & 0.6\\
		Deep Matting~\cite{xu2017deep} & 14.3 & 10.8 & 11 & 21 & 0.4 & 0.4 & 0.5 & 0.2 & 0.2 & \textbf{0.2} & \textbf{0.1} & \textbf{0.1} & 0.2 & 0.2 & 0.2 & 0.6 & \textbf{1.3} & \textbf{1.5} & \textbf{2.4} & \textbf{0.8} & \textbf{0.9} & 1.3 & \textbf{0.7} & \textbf{0.8} & \textbf{1.1} & \textbf{0.4} & \textbf{0.5} & 0.5\\
		\hline
	\end{tabular}
	%
%	\vspace{-8pt}
	\caption{Gradient errors (top 3) on the $ \tt alphamatting.com$ online benchmark.
	The lowest errors are boldfaced.}
	\label{tab:alphamatting.com}
	%
%	\vspace{-12pt}
\end{table*}

\begin{figure}[!tb]
	\captionsetup{font=small,singlelinecheck=true}
	\setlength{\abovecaptionskip}{10pt}
	\centering
	\includegraphics[width=3.2in,angle=0]{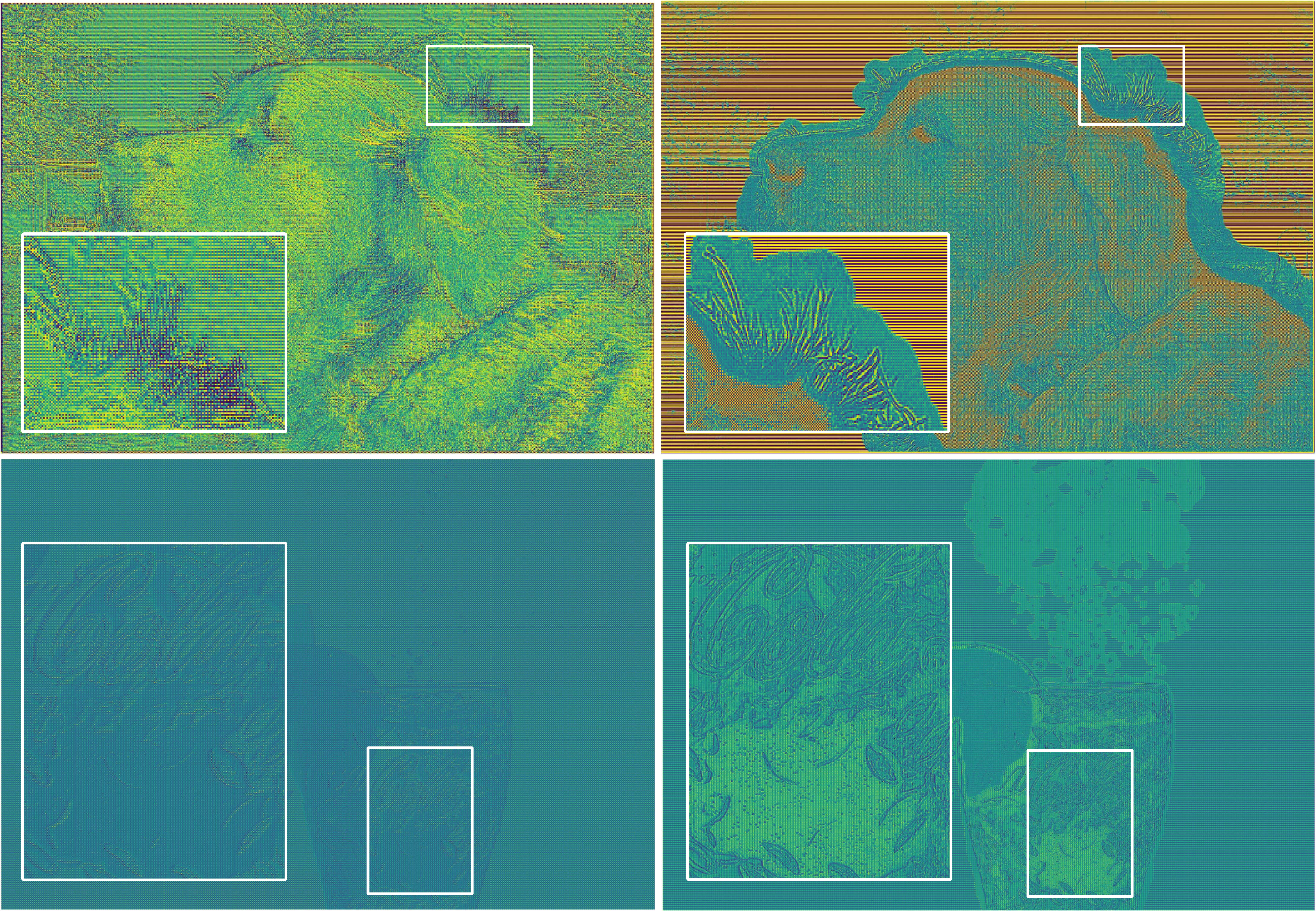}\vspace{-8pt}
	\caption{Visualization of the randomly initialized index map (left) and the learned index map (right) of HINs (top) and DINs (bottom). Best viewed by zooming in.}
	\label{fig:indices_visualization}
	\vspace{-12pt}
\end{figure}

\vspace{3pt}
\noindent\textbf{Index Map Visualization}. It is interesting to see what indices are learned by IndexNet. For the holistic index, the index map itself is a 2D matrix and is easily to be visualized. Regarding the depthwise index, we squeeze the index map along the channel dimension and calculate the average responses. Two examples of learned index maps are visualized in Fig.~\ref{fig:indices_visualization}. We observe that, initial random indices have poor delineation for edges, while learned indices automatically capture the complex structural and textual patterns, e.g., the fur of the dog, and even air bubbles in the water.

\subsection{alphamatting.com Online Benchmark}

\begin{figure*}[!tb]
	\captionsetup{font=small,singlelinecheck=true}
	\setlength{\abovecaptionskip}{10pt}
	\centering
	\includegraphics[width=.7\textwidth,angle=0]{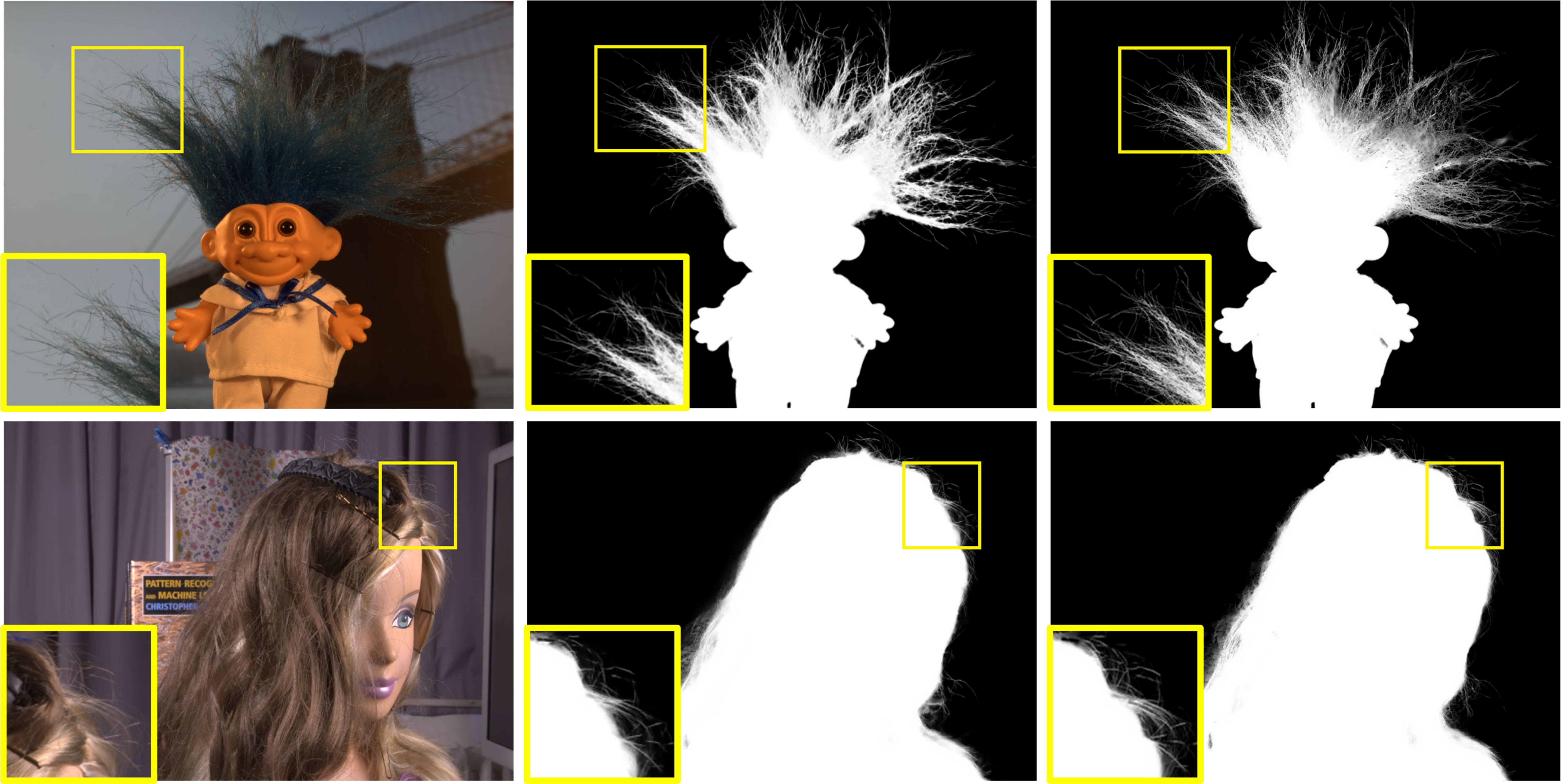}%\vspace{-8pt}
	\caption{Qualitative results on the $\tt alphamatting.com$ dataset. From left to right, the original image, deep image matting, ours.}
	\label{fig:alphamatting}
%	\vspace{-15pt}
\end{figure*}

We also report results on the $ \tt alphamatting.com $ online benchmark~\cite{rhemann2009perceptually}. We directly test our best model trained on the Adobe Image Dataset, without fine-tuning. Our approach (IndexNet Matting) ranks the first in terms of the gradient error among published methods, as shown in Table~\ref{tab:alphamatting.com}. According to the qualitative results in Fig.~\ref{fig:alphamatting}, our approach produces significantly better mattes on hair.

\subsection{Extensions to Other Visual Tasks}

We further evaluate IndexNet on other three visual tasks. For image classification, we compare three classification networks (LeNet~\cite{lecun1998gradient}, MobileNet~\cite{howard2017mobilenets} and VGG-16~\cite{simonyan2014very}) on the CIFAR-10 and CIFAR-100 datasets~\cite{krizhevsky2009learning} with/without IndexNet. For monocular depth estimation, we attach IndexNet upon a recent ResNet-50 based baseline~\cite{hu2019revisiting} and report the performance on the NYUDv2 dataset~\cite{silberman2012indoor}. On the task of scene understanding, we evaluate SegNet~\cite{badrinarayanan2017segnet} with/without IndexNet on the SUN-RGBD dataset~\cite{song2015sun}. Results show that IndexNet consistently improves the performance in all three tasks. We refer readers to the Supplement for quantitative and qualitative results.

\section{Conclusion}
Inspired by an observation in image matting, we delve deep into the role of indices and present an unified perspective of upsampling operators using the notion of index function. We show that an index function can be learned within a proposed index-guided encoder-decoder framework. In this framework, indices are learned with a flexible network module termed IndexNet, and are used to guide downsampling and upsampling using two operators called $\mathcal{IP}$ and $\mathcal{IU}$. IndexNet itself is also a sub-framework that can be designed depending on the task at hand. We instantiated, investigated three index networks, compared their conceptual differences, discussed their properties, and demonstrated their effectiveness on the task of image matting, image classification, depth prediction and scene understanding. We report state-of-the-art performance on image matting with a modified MobileNetv2-based model on the Composition-1k dataset. We believe that
IndexNet is an important step towards the design of generic upsampling operators.

Our model is simple with much room for improvement. %
It may be used as a strong baseline for future research. We plan to explore the applicability of IndexNet to other dense prediction tasks.

{\bf Acknowledgments}
We would like to thank Huawei Technologies for the donation of GPU cloud computing resources.

{\small
\bibliographystyle{ieee}
\bibliography{egbib}
}

\end{document}